\documentclass[10pt,twocolumn,letterpaper]{article}

\usepackage{iccv}
\usepackage{times}
\usepackage{epsfig}
\usepackage{graphicx}
\usepackage{amsmath}
\usepackage{amssymb}

\usepackage[caption=false,font=footnotesize]{subfig}
\usepackage{array}
\usepackage{booktabs}
\usepackage{xcolor}
\usepackage{color}
\usepackage{soul}
\usepackage{epstopdf}
\usepackage{tabu}
\usepackage[export]{adjustbox}
\usepackage{float}
\usepackage{enumitem}
\newcommand{\rr}{\textsuperscript{\tiny\textregistered}}

\makeatletter
\newcommand{\clearsubcaptcounter}{\setcounter{sub\@captype}{0}}
\makeatother

\newcommand*{\affaddr}[1]{#1} 

\usepackage[breaklinks=true,bookmarks=false]{hyperref}

\iccvfinalcopy 

\def\iccvPaperID{2329} 

\ificcvfinal\fi
\begin{document}

\title{DeepFuse: A Deep Unsupervised Approach for Exposure Fusion with Extreme Exposure Image Pairs} 
\author{%
K. Ram Prabhakar, V Sai Srikar, and R. Venkatesh Babu\\
\affaddr{Video Analytics Lab, Department of Computational and Data Sciences,}\\
\affaddr{Indian Institute of Science, Bangalore, India}\\
}
\maketitle
\thispagestyle{empty}

\begin{abstract}
We  present a novel deep learning architecture for fusing static multi-exposure images. Current multi-exposure fusion (MEF) approaches use hand-crafted features to fuse input sequence. However, the weak hand-crafted representations are not robust to varying input conditions. Moreover, they perform poorly for extreme exposure image pairs. Thus, it is highly desirable to have a method that is robust to varying input conditions and capable of handling extreme exposure without artifacts. Deep representations have known to be robust to input conditions and have shown phenomenal performance in a supervised setting. However, the stumbling block in using deep learning for MEF was the lack of sufficient training data and an oracle to provide the ground-truth for supervision. To address the above issues, we have gathered a large dataset of multi-exposure image stacks for training and to circumvent the need for ground truth images, we propose an unsupervised deep learning framework for MEF utilizing a no-reference quality metric as loss function. The proposed approach uses a novel CNN architecture trained to learn the fusion operation without reference ground truth image. The model fuses a set of common low level features extracted from each image to generate artifact-free perceptually pleasing results. We perform extensive quantitative and qualitative evaluation and show that the proposed technique outperforms existing state-of-the-art approaches for a variety of natural images.
\end{abstract}
\section{Introduction}
\begin{figure}[ht]
\centering
\mbox{\subfloat { \includegraphics[clip,trim=0cm 3.5cm 0cm 3.5cm ,width=8cm]{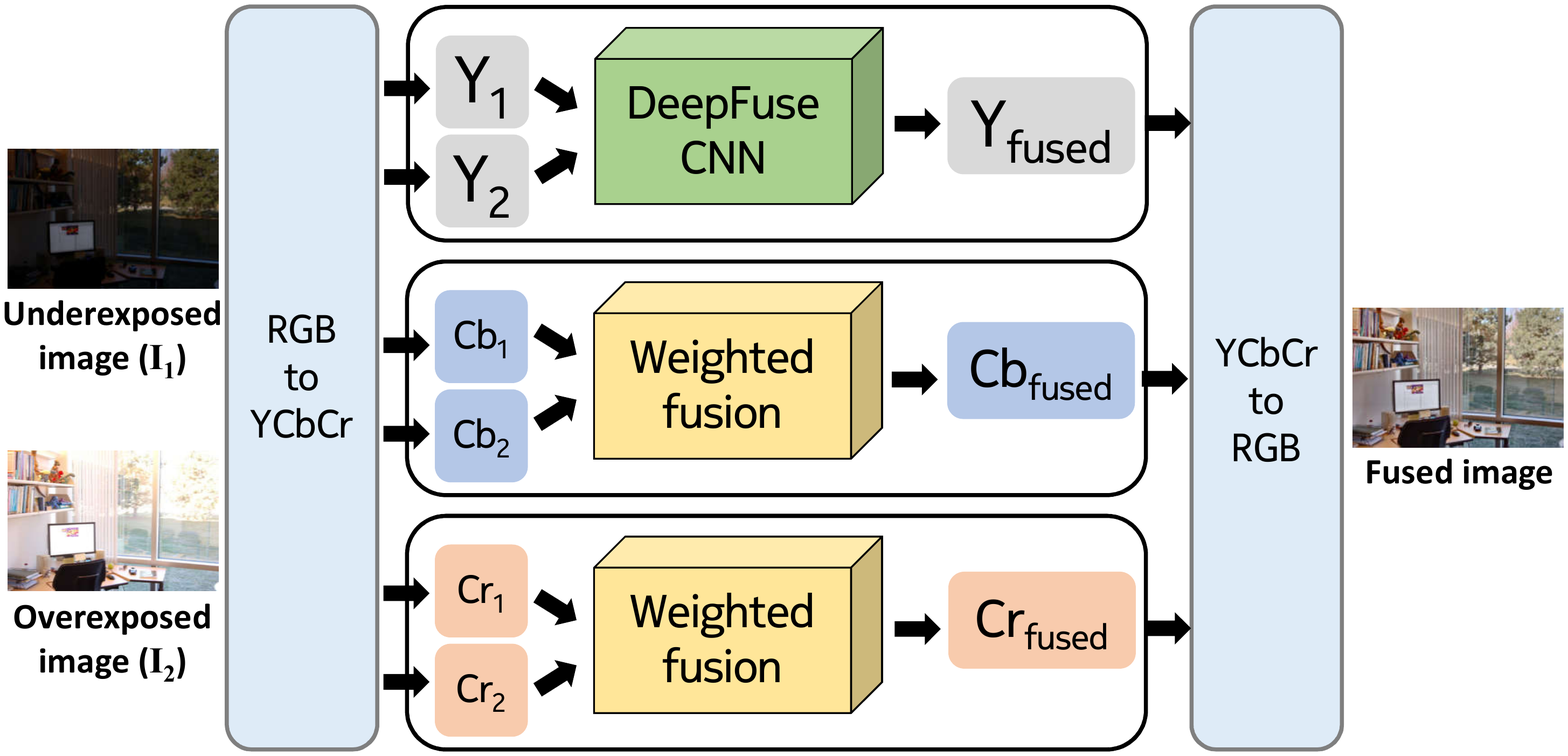}}}\hspace{0.6cm}%
\caption{Schematic diagram of the proposed method.}
\label{fig:overall}
\end{figure}
\vspace{-0.2cm}
High Dynamic Range Imaging (HDRI) is a photography technique that helps to capture better-looking photos in difficult lighting conditions. It helps to store all range of light (or brightness) that is perceivable by human eyes, instead of using limited range achieved by cameras. Due to this property, all objects in the scene look better and clear in HDRI, without being saturated (too dark or too bright) otherwise. 


The popular approach for HDR image generation is called as Multiple Exposure Fusion (MEF), in which, a set of static LDR images (further referred as exposure stack) with varying exposure is fused into a single HDR image. The proposed method falls under this category. Most of MEF algorithms work better when the exposure bias difference between each LDR images in exposure stack is minimum\footnote{Exposure bias value indicates the amount of exposure offset from the auto exposure setting of an camera. For example, EV 1 is equal to doubling auto exposure time (EV 0).}. Thus they require more LDR images (typically more than 2 images) in the exposure stack to capture whole dynamic range of the scene. It leads to more storage requirement, processing time and power. In principle, the long exposure image (image captured with high exposure time) has better colour and structure information in dark regions and short exposure image (image captured with less exposure time) has better colour and structure information in bright regions. Though fusing extreme exposure images is practically more appealing, it is quite challenging (existing approaches fail to maintain uniform luminance across image). Additionally, it should be noted that taking more pictures increases power, capture time and computational time requirements. Thus, we propose to work with exposure bracketed image pairs as input to our algorithm.

In this work, we present a data-driven learning method for fusing exposure bracketed static image pairs. To our knowledge this is the first work that uses deep CNN architecture for exposure fusion. The initial layers consists of a set of filters to extract common low-level features from each input image pair. These low-level features of input image pairs are fused for reconstructing the final result. The entire network is trained end-to-end using a no-reference image quality loss function. 

We train and test our model with a huge set of exposure stacks captured with diverse settings (indoor/outdoor, day/night, side-lighting/back-lighting, and so on). Furthermore, our model does not require parameter fine-tuning for varying input conditions. Through extensive experimental evaluations we demonstrate that the proposed architecture performs better than state-of-the-art approaches for a wide range of input scenarios.

\par The contributions of this work are as follows:
\begin{itemize}[noitemsep,topsep=0pt,parsep=0pt,partopsep=0pt]
\item A CNN based unsupervised image fusion algorithm for fusing exposure stacked static image pairs.
\item A new benchmark dataset that can be used for comparing various MEF methods.
\item An extensive experimental evaluation and comparison study against 7 state-of-the-art algorithms for variety of natural images. 
\end{itemize}
The paper is organized as follows. Section 2, we briefly review related works from literature. Section 3, we present our CNN based exposure fusion algorithm and discuss the details of experiments. Section 4, we provide the fusion examples and then conclude the paper with an insightful discussion in section 5.
\vspace{-0.1cm}
\section{Related Works}
\label{sec:related_works}
\vspace{-0.1cm}
Many algorithms have been proposed over the years for exposure fusion. However, the main idea remains the same in all the algorithms. The algorithms compute the weights for each image either locally or pixel wise.  The fused image would then be the weighted sum of the images in the input sequence. 	
\par Burt \emph{et al.} \cite{burt93} performed a Laplacian pyramid decomposition of the image and the weights are computed using local energy and correlation between the pyramids. Use of Laplacian pyramids reduces the chance of unnecessary artifacts. Goshtasby \emph{et al.} \cite{Gosh05} take non-overlapping blocks with highest information from each image to obtain the fused result. This is prone to suffer from block artifacts. Mertens \emph{et al.} \cite{mertens2007exposure} perform exposure fusion using simple quality metrics such as contrast and saturation. However, this suffers from hallucinated edges and mismatched color artifacts.  

\par Algorithms which make use of edge preserving filters like Bilateral filters are proposed in \cite{raman2009bilateral}. As this does not account for the luminance of the images, the fused image has dark region leading to poor results. A gradient based approach to assign the weight was put forward by Zhang \emph{et al.} \cite{zhang2012reference}. In a series of papers by Li \emph{et al.} \cite{shutao12}, \cite{shutao13} different approaches to exposure fusion have been reported. In their early works they solve a quadratic optimization to extract finer details and fuse them. In one of their later works \cite{shutao13}, they propose a Guided Filter based approach. 

\begin{figure*}[ht]
\centering
\mbox{\subfloat { \includegraphics[clip,trim=0cm 5cm 0cm 6cm , height=5cm]{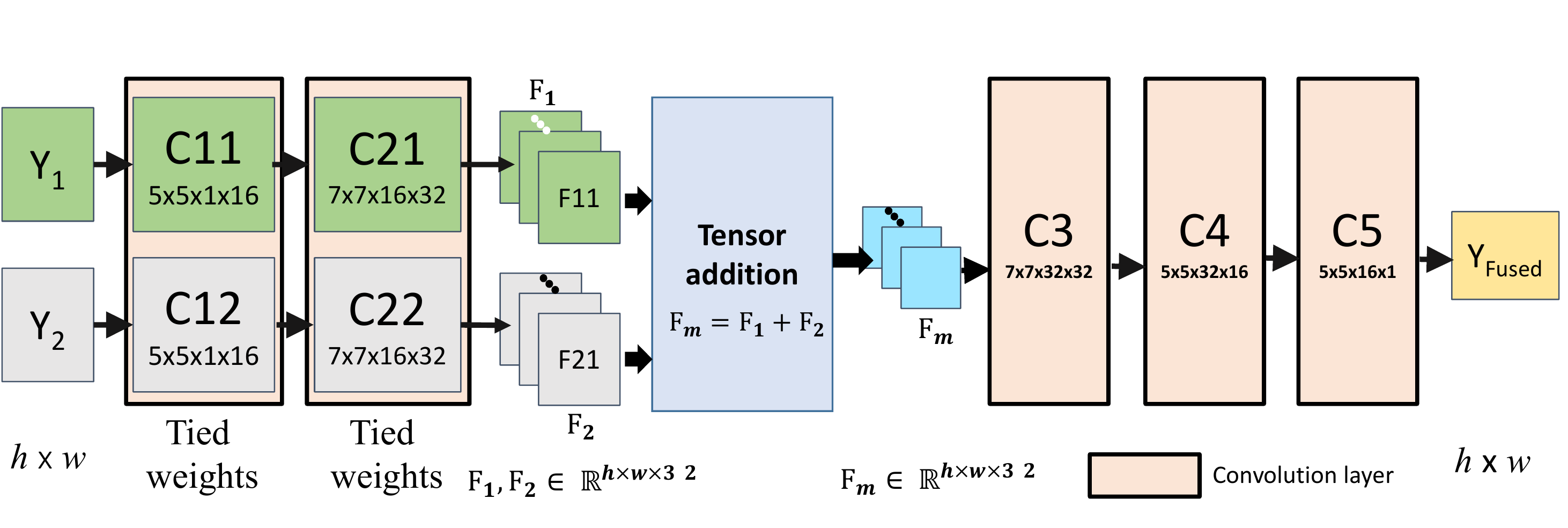}}}
\caption{Architecture of proposed image fusion CNN illustrated for input exposure stack with images of size $h\times w$. The pre-fusion layers C1 and C2 that share same weights, extract low-level features from input images. The feature pairs of input images are fused into a single feature by merge layer. The fused features are input to reconstruction layers to generate fused image $Y_{fused} $.}
\label{fig:arch_types}
\end{figure*}
Shen \emph{et al.} \cite{shen2014exposure} proposed a fusion technique using quality metrics such as local contrast and color consistency. The random walk approach they perform gives a global optimum solution to the fusion problem set in a probabilistic fashion. 
\par All of the above works rely on hand-crafted features for image fusion. These methods are not robust in the sense that the parameters need to be varied for different input conditions say, linear and non-linear exposures, filter size depends on image sizes. To circumvent this parameter tuning we propose a feature learning based approach using CNN. In this work we learn suitable features for fusing exposure bracketed images. Recently, Convolutional Neural Network (CNN) have shown impressive performance across various computer vision tasks \cite{lecun2015deep}. While CNNs have produced state-of-the-art results in many high-level computer vision tasks like recognition (\cite{he2016deep}, \cite{sarvadevabhatla2016enabling}), object detection \cite{li2016r}, Segmentation \cite{he2017mask}, semantic labelling \cite{pinheiro2013recurrent}, visual question answering \cite{antol2015vqa} and much more, their performance on low-level image processing problems such as filtering \cite{nithish2017} and fusion \cite{prabhakar2016ghosting} is not studied extensively. In this work we explore the effectiveness of CNN for the task of multi-exposure image fusion. 
\par To our knowledge, use of CNNs for multi-exposure fusion is not reported in literature. The other machine learning approach is based on a regression method called Extreme Learning Machine (ELM) \cite{wang2012extreme}, that feed saturation level, exposedness, and contrast into the regressor to estimate the importance of each pixel. Instead of using hand crafted features, we use the data to learn a representation right from the raw pixels. 

\vspace{-0.1cm}
\section{Proposed Method}
\label{sec:proposed_method}
\vspace{-0.1cm}
In this work, we propose an image fusion framework using CNNs. Within a span of couple years, Convolutional Neural Networks have shown significant success in high-end computer vision tasks. They are shown to learn complex mappings between input and output with the help of sufficient training data. CNN learns the model parameters by optimizing a loss function in order to predict the result as close as to the ground-truth. For example, let us assume that input \textbf{x} is mapped to output \textbf{y} by some complex transformation \emph{f}. The CNN can be trained to estimate the function \emph{f} that minimizes the difference between the expected output \textbf{y} and obtained output \( \hat{\textbf{y}} \). The distance between \textbf{y} and \( \hat{\textbf{y}} \) is calculated using a loss function, such as mean squared error function. Minimizing this loss function leads to better estimate of required mapping function. 

Let us denote the input exposure sequence and fusion operator as $I$ and $O(I)$. The input images are assumed to be registered and aligned using existing registration algorithms, thus avoiding camera and object motion. We model $O(I)$ with a feed-forward process $F_W(I)$. Here, $F$ denotes the network architecture and $W$ denotes the weights learned by minimizing the loss function. As the expected output $O(I)$ is absent for MEF problem, the squared error loss or any other full reference error metric cannot be used. Instead, we make use of no-reference image quality metric MEF SSIM proposed by Ma \textit{et al.} \cite{ma2015perceptual} as loss function. MEF SSIM is based on structural similarity index metric (SSIM) framework \cite{wang2004image}. It makes use of statistics of a patch around individual pixels from input image sequence to compare with result. It measures the loss of structural integrity as well as luminance consistency in multiple scales (see section \ref{subsec:loss_fun} for more details). 

An overall scheme of proposed method is shown in Fig. \ref{fig:overall}. The input exposure stack is converted into YCbCr color channel data. The CNN is used to fuse the luminance channel of the input images. This is due to the fact that the image structural details are present in luminance channel and the brightness variation is prominent in luminance channel than chrominance channels. The obtained luminance channel is combined with chroma (Cb and Cr) channels generated using method described in section \ref{subsec:chrom}. The following subsection details the network architecture, loss function and the training procedure.
\vspace{-0.2cm}
\subsection{DeepFuse CNN}
\vspace{-0.2cm}
\par The learning ability of CNN is heavily influenced by right choice of architecture and loss function. A simple and naive architecture is to have a series of convolutional layers connected in sequential manner. The input to this architecture would be exposure image pairs stacked in third dimension. Since the fusion happens in the pixel domain itself, this type of architecture does not make use of feature learning ability of CNNs to a great extent.
\par
\begin{figure}[t]
\centering
\includegraphics[width=3in]{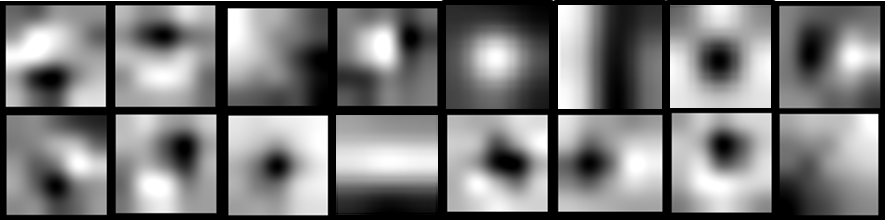}
\caption{\textbf{Filter Visualization.} Some of the filters learnt in first layer resemble Gaussian, Difference of Gaussian and Laplacian of Gaussian filters. Best viewed electronically, zoomed in.}
\label{fig_layerweights}
\end{figure}

 The proposed network architecture for image fusion is illustrated in Fig. \ref{fig:arch_types}. The proposed architecture has three components: feature extraction layers, a fusion layer and reconstruction layers. As shown in Fig. \ref{fig:arch_types}, the under-exposed and the over-exposed images ($Y_1$ and $Y_2$) are input to separate channels (channel 1 consists of C11 and C21 and channel 2 consists of C12 and C22). The first layer (C11 and C12) contains 5 $\times $ 5 filters to extract low-level features such as edges and corners. The weights of pre-fusion channels are \emph{tied}, C11 and C12 (C21 and C22) share same weights. The advantage of this architecture is three fold: first, we force the network to learn the same features for the input pair. That is, the F11 and F21 are same feature type. Hence, we can simply combine the respective feature maps via fusion layer. Meaning, the first feature map of image 1 (F11) and the first feature map of image 2 (F21) are added and this process is applied for remaining feature maps as well. Also, adding the features resulted in better performance than other choices of combining features (see Table \ref{tab2}). In feature addition, similar feature types from both images are fused together. Optionally one can choose to concatenate features, by doing so, the network has to figure out the weights to merge them. In our experiments, we observed that feature concatenation can also achieve similar results by increasing the number of training iterations, increasing number of filters and layers after C3. This is understandable as the network needs more number of iterations to figure out appropriate fusion weights. In this tied-weights setting, we are enforcing the network to learn filters that are invariant to brightness changes. This is observed by visualizing the learned filters (see Fig. \ref{fig_layerweights}). In case of tied weights, few high activation filters have center surround receptive fields (typically observed in retina). These filters have learned to remove the mean from neighbourhood, thus effectively making the features brightness invariant. Second, the number of learnable filters is reduced by half. Third, as the network has low number of parameters, it converges quickly. The obtained features from C21 and C22 are fused by merge layer. The result of fuse layer is then passed through another set of convolutional layers (C3, C4 and C5) to reconstruct final result ($Y_{fused}$) from fused features.
\vspace{-0.5cm}
\subsubsection{MEF SSIM loss function}
\label{subsec:loss_fun}
\vspace{-0.2cm}
\par In this section, we will discuss on computing loss without using reference image by MEF SSIM image quality measure \cite{ma2015perceptual}. Let $\{{\mathbf{y}}_k\}$=$\{{\mathbf{y}}_k|k$=1,2$\}$ denote the image patches extracted at a pixel location $p$ from input image pairs and ${\mathbf{y}}_{f}$ denote the patch extracted from CNN output fused image at same location $p$. The objective is to compute a score to define the fusion performance given ${\mathbf{y}}_k$ input patches and ${\mathbf{y}}_f$ fused image patch. 

In SSIM \cite{wang2004image} framework, any patch can be modelled using three components: structure (${\mathbf{s}}$), luminance ($l$) and contrast ($c$). The given patch is decomposed into these three components as:
\begin{align} 
{\mathbf{y}}_{k}=&\|{\mathbf{y}}_{k} - \mu _{{\mathbf{y}}_{k}}\| \cdot \frac {{\mathbf{y}}_{k} - \mu _{{\mathbf{y}}_{k}}}{\|{\mathbf{y}}_{k} - \mu _{{\mathbf{y}}_{k}}\|} + \mu _{{\mathbf{y}}_{k}} \notag \\
=&\|\tilde {\mathbf{y}}_{k}\| \cdot \frac {\tilde {\mathbf{y}}_{k}}{\|\tilde {\mathbf{y}}_{k}\|} + \mu _{{\mathbf{y}}_{k}} \notag \\
=&c_{k} \cdot {\mathbf{s}}_{k} + l_{k}, 
\end{align}
where, $\parallel\cdot\parallel$ is the $\ell_2$ norm of patch, $\mu _{{\mathbf{y}}_{k}}$ is the mean value of ${\mathbf{y}}_{k}$ and $\tilde {\mathbf{y}}_{k}$ is the mean subtracted patch. As the higher contrast value means better image, the desired contrast value ($\hat{c}$) of the result is taken as the highest contrast value of $\{c_k\}$, (i.e.) \[\hat{c} = \underset{\{k=1,2\}}{\max } c_{k}\] 
The structure of the desired result ($\hat{{\mathbf{s}}}$) is obtained by weighted sum of structures of input patches as follows,
\begin{equation} 
\bar {\mathbf{s}} = \frac {\sum _{k = 1}^{2}w\left ({\tilde {\mathbf{y}}_{k}}\right ){\mathbf{s}}_{k}}{\sum _{k = 1}^{2}w\left ({\tilde {\mathbf{y}}_{k}}\right )} \quad {\rm and} \quad \hat {\mathbf{s}} = \frac {\bar {\mathbf{s}}}{\|\bar {\mathbf{s}}\|}, 
\end{equation}

where the weighting function assigns weight based on structural consistency between input patches. The weighting function assigns equal weights to patches, when they have dissimilar structural components. In the other case, when all input patches have similar structures, the patch with high contrast is given more weight as it is more robust to distortions. The estimated $\hat{s}$ and $\hat{c}$ is combined to produce desired result patch as,
\begin{equation}
\hat{{\mathbf{y}}} = \hat{c} \cdot \hat{s}
\end{equation}
As the luminance comparison in the local patches is insignificant, the luminance component is discarded from above equation. Comparing luminance at lower spatial resolution does not reflect the global brightness consistency. Instead, performing this operation at multiple scales would effectively capture global luminance consistency in coarser scale and local structural changes in finer scales. The final image quality score for pixel $p$ is calculated using SSIM framework,


\begin{equation} 
Score(p) = \frac {2\sigma _{\hat {\mathbf{y}}{\mathbf{y}}_f} + C}{{\sigma ^{2}_{\hat {\mathbf{y}}}} + \sigma^{2}_{{\mathbf{y}}_f} + C}, 
\end{equation}
where, $\sigma^2_{\hat{{\mathbf{y}}}}$ is variance and $\sigma _{\hat {\mathbf{y}}{\mathbf{y}}_f}$ is covariance between $\hat{{\mathbf{y}}}$ and ${\mathbf{y}}_f$. The total loss is calculated as,
\begin{equation}
Loss = 1 - \frac{1}{N}\sum_{p\in P} Score(p)
\end{equation} 
where $N$ is the total number of pixels in image and $P$ is the set of all pixels in input image. The computed loss is backpropagated to train the network. The better performance of MEF SSIM is attributed to its objective function that maximizes structural consistency between fused image and each of input images. 
\begin{table}[t]
\centering
\caption{\textbf{Choice of blending operators}: Average MEF SSIM scores of 23 test images generated by CNNs trained with different feature blending operations. The maximum score is highlighted in bold. Results illustrate that adding the feature tensors yield better performance. Results by addition and mean methods are similar, as both operations are very similar, except for a scaling factor. Refer text for more details.}
\label{tab2}
\begin{tabular}{@{}ccccc@{}}
\toprule
Product & Concatenation & Max    & Mean   & Addition \\ \midrule
0.8210  & 0.9430        & 0.9638 & 0.9750 & \textbf{0.9782}   \\ \bottomrule
\end{tabular}
\end{table}
\vspace{-0.2cm}
\subsection{Training}
\begin{figure*}[ht]
\centering
\mbox{\subfloat[Underexposed image] { \includegraphics[width =2.75cm]{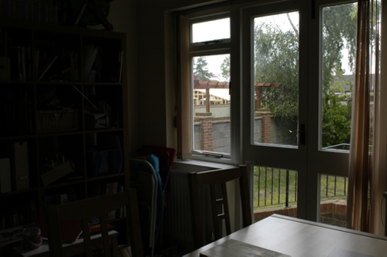}}}\hspace{0.001em}%
\mbox{\subfloat[Overexposed image] { \includegraphics[width =2.75cm]{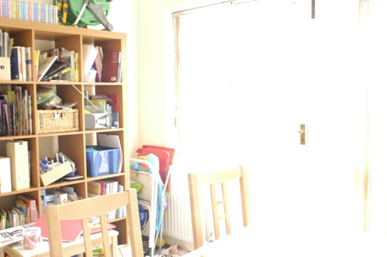}}}\hspace{0.001em}%
\mbox{\subfloat[Li \emph{et al.} \cite{shutao12}] { \includegraphics[width =2.75cm]{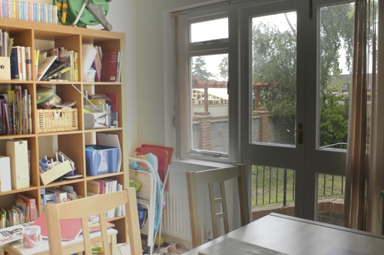}}}\hspace{0.001em}%
\mbox{\subfloat[Li \emph{et al.} \cite{shutao13}] { \includegraphics[width =2.75cm]{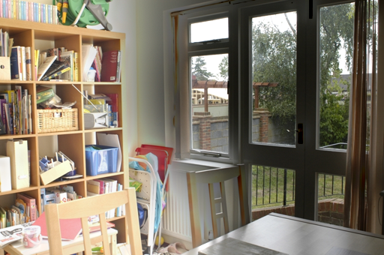}}}\hspace{0.001em}%
\mbox{\subfloat[Mertens \emph{et al.} \cite{mertens2007exposure}] { \includegraphics[width =2.75cm]{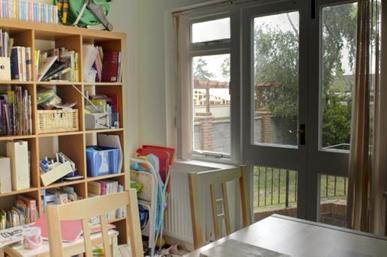}}}\hspace{0.001em}%
\mbox{\subfloat[Raman \emph{et al.} \cite{Shanmuga2011}] { \includegraphics[width =2.75cm]{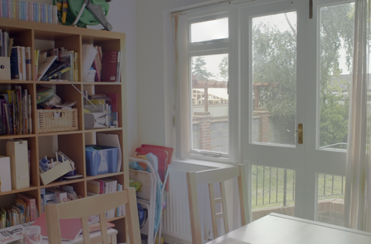}}}\hspace{0.002em}%

\mbox{\subfloat[Shen \emph{et al.} \cite{shen2011generalized}] { \includegraphics[width =2.75cm]{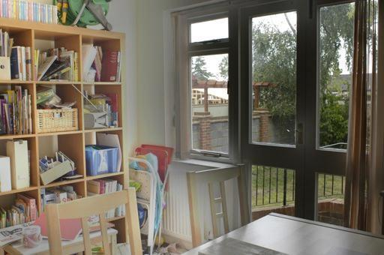}}}\hspace{0.002em}%
\mbox{\subfloat[Ma \emph{et al.} \cite{ma2015multi}] { \includegraphics[width =2.75cm]{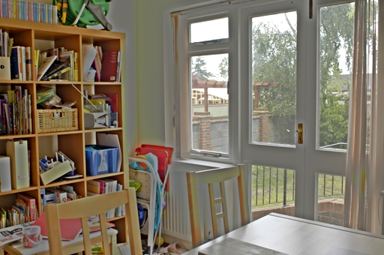}}}\hspace{0.002em}%
\mbox{\subfloat[Guo \emph{et al.} \cite{zhengguo2017detail}] { \includegraphics[width =2.75cm]{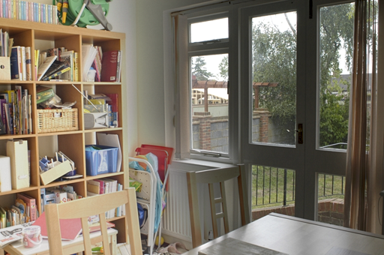}}}\hspace{0.002em}%
\mbox{\subfloat[DF-Baseline] { \includegraphics[width =2.75cm]{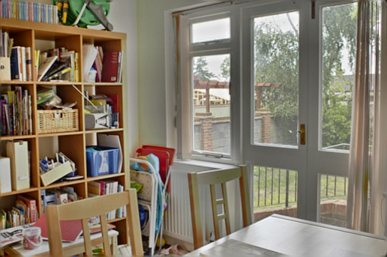}}}\hspace{0.002em}%
\mbox{\subfloat[DF-Unsupervised] { \includegraphics[width =2.75cm]{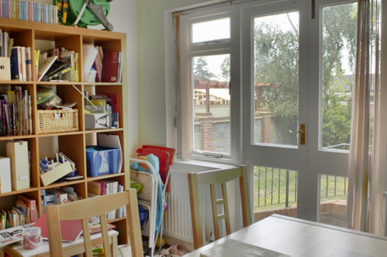}}}\hspace{0.002em}%
\caption{Results for House image sequence. Image courtesy of Kede ma. Best viewed in color.}
\label{fig:lighthouse}
\end{figure*}
\vspace{-0.2cm}
We have collected 25 exposure stacks that are available publicly \cite{EMPADataset}. In addition to that, we have curated 50 exposure stacks with different scene characteristics. The images were taken with standard camera setup and tripod. Each scene consists of 2 low dynamic range images with $ \pm2 $ EV difference. The input sequences are resized to 1200 $ \times $ 800 dimensions. We give priority to cover both indoor and outdoor scenes. From these input sequences, 30000 patches of size 64 $ \times$64 were cropped for training. We set the learning rate to $10^{-4} $ and train the network for 100 epochs with all the training patches being processed in each epoch.
\vspace{-0.2cm}
\subsection{Testing}
\label{subsec:chrom}
\vspace{-0.2cm}
We follow the standard cross-validation procedure to train our model and test the final model on a disjoint test set to avoid over-fitting. While testing, the trained CNN takes the test image sequence and generates the luminance channel ($Y_{fused}$) of fused image. The chrominance components of fused image, $Cb_{fused}$ and $Cr_{fused}$, are obtained by weighted sum of input chrominance channel values. 

The crucial structural details of the image tend to be present mainly in $Y$ channel. Thus, different fusion strategies are followed in literature for $Y$ and $Cb$/$Cr$ fusion (\cite{prabhakar2016ghosting}, \cite{tico2009image}, \cite{wang2009exposure}). Moreover, MEF SSIM loss is formulated to compute the score between 2 gray-scale ($Y$) images. Thus, measuring MEF SSIM for $Cb$ and $Cr$ channels may not be meaningful. Alternately, one can choose to fuse RGB channels separately using different networks. However, there is typically a large correlation between RGB channels. Fusing RGB independently fails to capture this correlation and introduces noticeable color difference. Also, MEF-SSIM is not designed for RGB channels. Another alternative is to regress RGB values in a single network, then convert them to a $Y$ image and compute MEF SSIM loss. Here, the network can focus more on improving $Y$ channel, giving less importance to color. However, we observed spurious colors in output which were not originally present in input.

We follow the procedure used by Prabhakar \textit{et al.} \cite{prabhakar2016ghosting} for chrominance channel fusion. If $x_1$ and $x_2$ denote the $Cb$ (or $Cr$) channel value at any pixel location for image pairs, then the fused chrominance value $x$ is obtained as follows,
	\begin{equation}
	x = \dfrac{x_1 (|x_1 - \tau|) + x_2 (|x_2 - \tau|)}{ |x_1 - \tau| + |x_2 - \tau|}
	\label{eqn:chrom}
	\end{equation}
The fused chrominance value is obtained by weighing two chrominance values with $\tau$ subtracted value from itself. The value of $\tau$ is chosen as 128. The intuition behind this approach is to give more weight for good color components and less for saturated color values. The final result is obtained by converting \{$Y_{fused}$, $Cb_{fused}$, $Cr_{fused}$\} channels into RGB image. 
\begin{figure*}[!ht]
\centering
\subfloat[Underexposed input]{\includegraphics[width=1.1in]{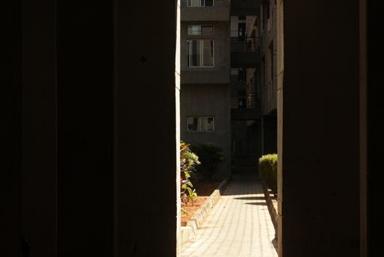}%
\label{fig1_ue_case}}
\hspace{0.01cm}
\subfloat[Overexposed input]{\includegraphics[width=1.1in]{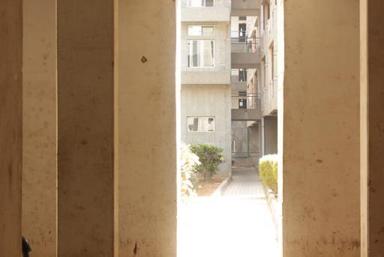}%
\label{fig1_ue_case}}
\hspace{0.01cm}
\subfloat[Mertens \emph{et al.} \cite{mertens2007exposure}]{\includegraphics[width=1.1in]{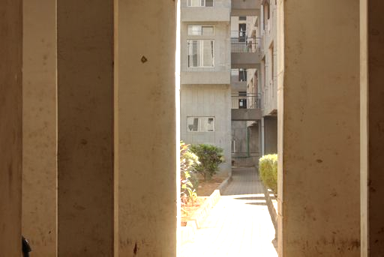}%
\label{fig1_s24_mer}}
\hspace{0.01cm}
\subfloat[Zoomed result of (c)]{\includegraphics[width=1.15in]{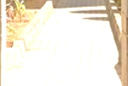}%
\label{fig1_s24_mer}}
\hspace{0.01cm}
\subfloat[DF - Unsupervised]{\includegraphics[width=1.1in]{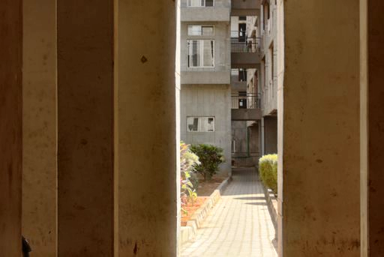}%
\label{fig1_s24_mer}}
\hspace{0.01cm}
\subfloat[Zoomed result of (e)]{\includegraphics[width=1.1in]{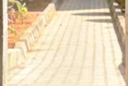}%
\label{fig1_s24_mer}}
\\	
\subfloat[Underexposed input]{\includegraphics[width=1.25in]{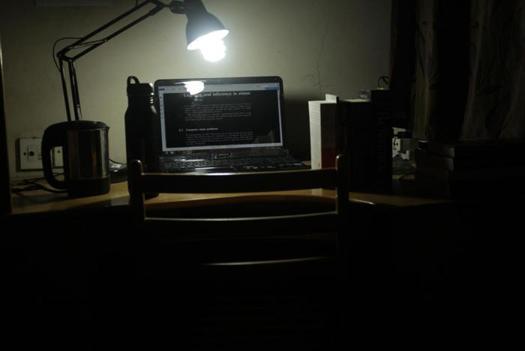}%
\label{fig1_ue_case}}
\hspace{0.01cm}
\subfloat[Overexposed input]{\includegraphics[width=1.25in]{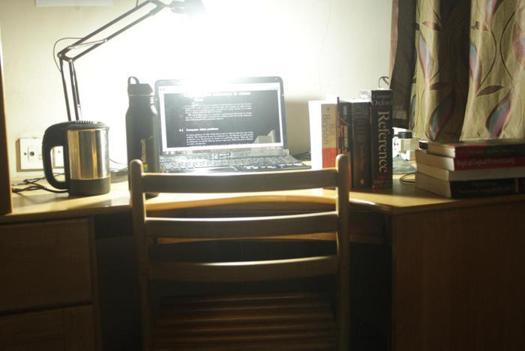}%
\label{fig1_ue_case}}
\hspace{0.01cm}
\subfloat[Mertens \emph{et al.} \cite{mertens2007exposure}]{\includegraphics[width=1.25in]{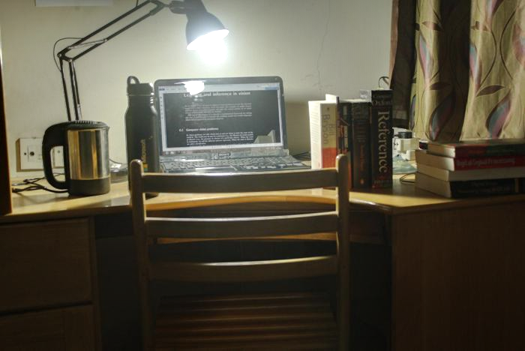}%
\label{fig1_s24_mer}}
\hspace{0.01cm}
\subfloat[Zoomed result of (i)]{\includegraphics[width=0.82in]{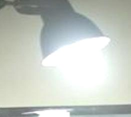}%
\label{fig1_s24_mer}}
\hspace{0.01cm}
\subfloat[DF - Unsupervised]{\includegraphics[width=1.25in]{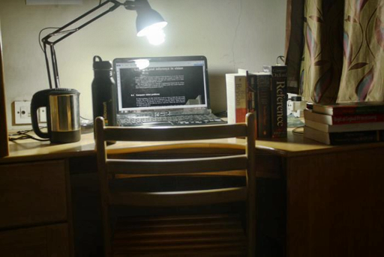}%
\label{fig1_s24_mer}}
\hspace{0.01cm}
\subfloat[Zoomed result of (k)]{\includegraphics[width=0.82in]{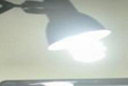}%
\label{fig1_s24_mer}}
\hspace{0.01cm}
\caption{Comparison of the proposed method with Mertens \emph{et al.} \cite{mertens2007exposure}. The Zoomed region of the result by Mertens \emph{et al.} in (d) show that some highlight regions are not completely retained from input. The zoomed region of the result by Mertens \emph{et al.} in (j) show that fine details of lamp are missing.}
\label{fig_hostel1}
\end{figure*}

\begin{table*}[th]
\tiny{
\centering
\caption{MEF SSIM scores of different methods against DeepFuse (DF) for test images. Bolded values indicate the highest score by that corresponding column algorithm than others for that row image sequence.}
\label{tab:mef_ssim}
\resizebox{\textwidth}{!}{%
\begin{tabular}{|l|c|c|c|c|c|c|c|c|c|}
\hline
\textit{} & \textbf{{ \textit{Mertens09}}} & \textbf{{ \textit{Raman11}}} & \textbf{{ \textit{Li12}}} & \textbf{ \textit{Li13}} & \textbf{ \textit{Shen11}} & \textbf{ \textit{Ma15}} & \textbf{ \textit{Guo17}} & \textbf{ \textit{DF-Baseline}} & \textbf{ \textit{DF-UnSupervised}} \\ \hline \hline
\textit{AgiaGalini} & 0.9721 & 0.9343 & 0.9438 & 0.9409 & 0.8932 & 0.9465 & 0.9492 & 0.9477 & \textbf{0.9813} \\ \hline
\textit{Balloons} & 0.9601 & 0.897 & 0.9464 & 0.9366 & 0.9252 & 0.9608 & 0.9348& 0.9717 & \textbf{0.9766} \\ \hline
\textit{Belgium house} & 0.9655 & 0.8924 & 0.9637 & 0.9673 & 0.9442 & 0.9643 & 0.9706&0.9677 & \textbf{0.9727} \\ \hline
\textit{Building} & 0.9801 & 0.953 & 0.9702 & 0.9685 & 0.9513 & 0.9774 & 0.9666&0.965 & \textbf{0.9826} \\ \hline
\textit{Cadik lamp} & 0.9658 & 0.8696 & 0.9472 & 0.9434 & 0.9152 & 0.9464 &0.9484& \textbf{0.9683} & 0.9638 \\ \hline
\textit{Candle} & 0.9681 & 0.9391 & 0.9479 & 0.9017 & 0.9441 & 0.9519 & 0.9451&0.9704 & \textbf{0.9893} \\ \hline
\textit{Chinese garden} & \textbf{0.990} & 0.8887 & 0.9814 & 0.9887 & 0.9667 & \textbf{0.990} & 0.9860&0.9673 & 0.9838 \\ \hline
\textit{Corridor} & 0.9616 & 0.898 & 0.9709 & 0.9708 & 0.9452 & 0.9592 &0.9715& \textbf{0.9740} & \textbf{0.9740} \\ \hline
\textit{Garden} & 0.9715 & 0.9538 & 0.9431 & 0.932 & 0.9136 & 0.9667 &0.9481& 0.9385 & \textbf{0.9872} \\ \hline
\textit{Hostel} & 0.9678 & 0.9321 & 0.9745 & 0.9742 & 0.9649 & 0.9712 &0.9757& 0.9715 & \textbf{0.985} \\ \hline
\textit{House} & \textbf{0.9748} & 0.8319 & 0.9575 & 0.9556 & 0.9356 & 0.9365 &0.9623& 0.9601 & 0.9607 \\ \hline
\textit{Kluki Bartlomiej} & \textbf{0.9811} & 0.9042 & 0.9659 & 0.9645 & 0.9216 & 0.9622 &0.9680& 0.9723 & 0.9742 \\ \hline
\textit{Landscape} & 0.9778 & 0.9902 & 0.9577 & 0.943 & 0.9385 & 0.9817 &0.9467& 0.9522 & \textbf{0.9913} \\ \hline
\textit{Lighthouse} & 0.9783 & 0.9654 & 0.9658 & 0.9545 & 0.938 & 0.9702 &0.9657& 0.9728 & \textbf{0.9875} \\ \hline
\textit{Madison capitol} & 0.9731 & 0.8702 & 0.9516 & 0.9668 & 0.9414 & 0.9745 &0.9711& 0.9459 & \textbf{0.9749} \\ \hline
\textit{Memorial} & 0.9676 & 0.7728 & 0.9644 & \textbf{0.9771} & 0.9547 & 0.9754 &0.9739& 0.9727 & 0.9715 \\ \hline
\textit{Office} & \textbf{0.9749} & 0.922 & 0.9367 & 0.9495 & 0.922 & 0.9746 &0.9624& 0.9277 & \textbf{0.9749} \\ \hline
\textit{Room} & 0.9645 & 0.8819 & 0.9708 & \textbf{0.9775} & 0.9543 & 0.9641 &0.9725& 0.9767 & 0.9724 \\ \hline
\textit{SwissSunset} & 0.9623 & 0.9168 & 0.9407 & 0.9137 & 0.8155 & 0.9512 &0.9274& 0.9736 & \textbf{0.9753} \\ \hline
\textit{Table} & 0.9803 & 0.9396 & 0.968 & 0.9501 & 0.9641 & 0.9735 &0.9750& 0.9468 & \textbf{0.9853} \\ \hline
\textit{TestChart1} & 0.9769 & 0.9281 & 0.9649 & 0.942 & 0.9462 & 0.9529 &0.9617& 0.9802 & \textbf{0.9831} \\ \hline
\textit{Tower} & \textbf{0.9786} & 0.9128 & 0.9733 & 0.9779 & 0.9458 & 0.9704 &0.9772& 0.9734 & 0.9738 \\ \hline
\textit{Venice} & \textbf{0.9833} & 0.9581 & 0.961 & 0.9608 & 0.9307 & 0.9836 &0.9632& 0.9562 & 0.9787 \\ \hline 
\end{tabular}%
}}
\end{table*}

\begin{figure*}[ht]
\centering
\subfloat{\includegraphics[width=1.2in]{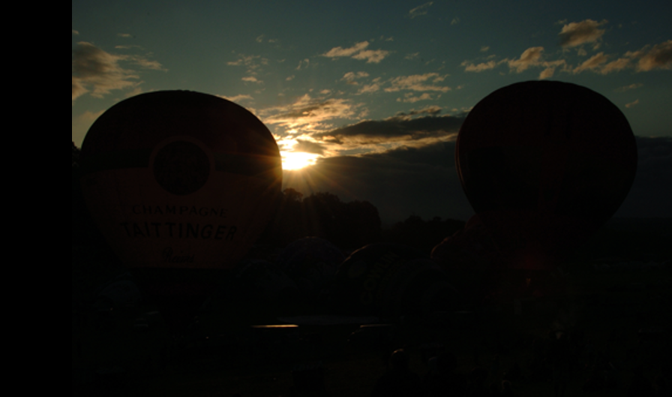}%
\label{fig1_ue_case}}
\hspace{0.05cm}
\subfloat{\includegraphics[width=1.05in]{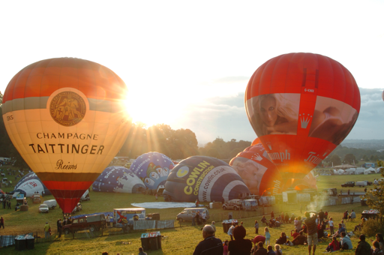}%
\label{fig1_ue_case}}
\hspace{0.05cm}
\subfloat{\includegraphics[width=1.05in]{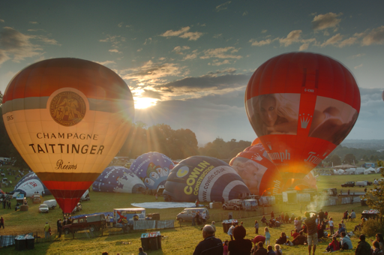}%
\label{fig1_s24_mer}}
\hspace{0.05cm}
\subfloat{\includegraphics[width=1.05in]{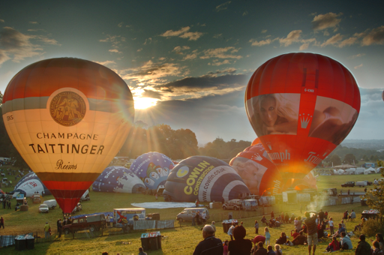}%
\label{fig1_s24_mer}}
\hspace{0.05cm}
\subfloat{\includegraphics[width=1.05in]{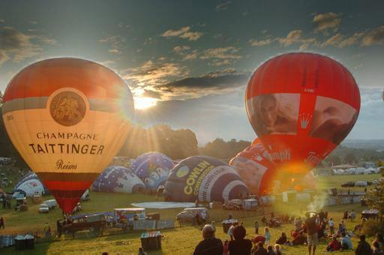}%
\label{fig1_s24_mer}}
\hspace{0.05cm}
\subfloat{\includegraphics[width=1.05in]{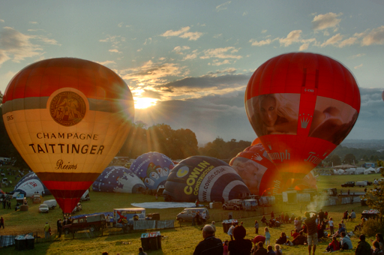}%
\label{fig1_s24_mer}}
\\\vspace{-1.5mm}
\clearsubcaptcounter
\subfloat[UE input]{\includegraphics[width=1.2in]{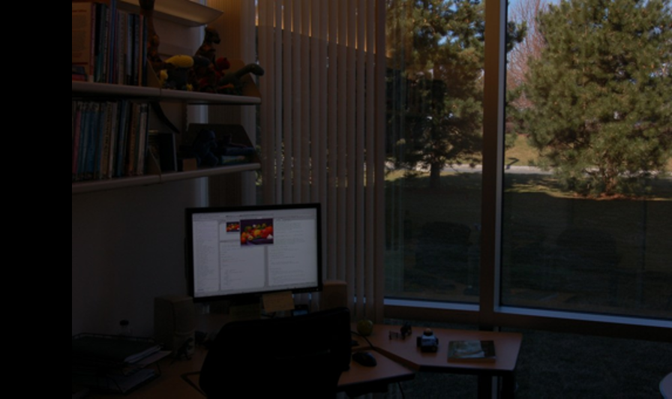}%
\label{fig1_ue_case}}
\hspace{0.05cm}
\subfloat[OE input]{\includegraphics[width=1.05in]{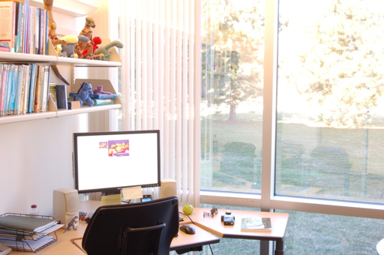}%
\label{fig1_ue_case}}
\hspace{0.05cm}
\subfloat[Li \emph{et al.} \cite{shutao12}]{\includegraphics[width=1.05in]{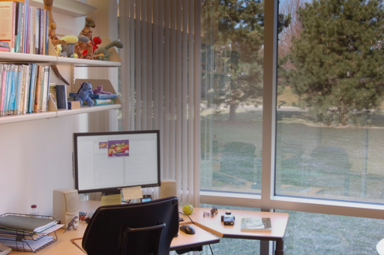}%
\label{fig1_s24_mer}}
\hspace{0.05cm}
\subfloat[Li \emph{et al.} \cite{shutao13}]{\includegraphics[width=1.05in]{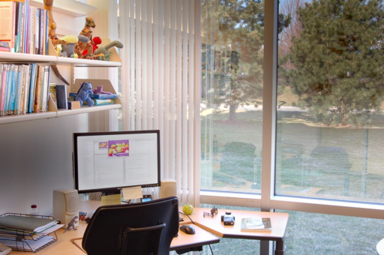}%
\label{fig1_s24_mer}}
\hspace{0.05cm}
\subfloat[Shen \emph{et al.} \cite{shen2011generalized}]{\includegraphics[width=1.05in]{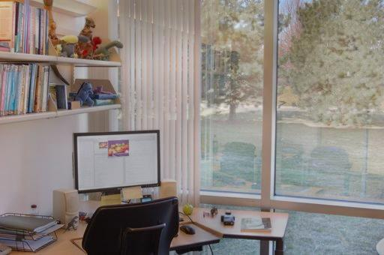}%
\label{fig1_s24_mer}}
\hspace{0.05cm}
\subfloat[DeepFuse]{\includegraphics[width=1.05in]{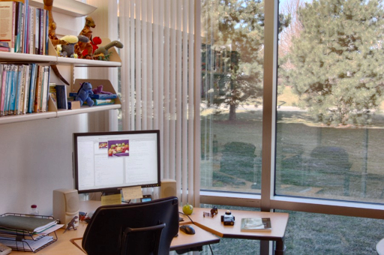}%
\label{fig1_s24_mer}}
\caption{Comparison of the proposed method with Li \emph{et al.} \cite{shutao12}, Li \emph{et al.} \cite{shutao13} and Shen \emph{et al.} \cite{shen2011generalized} for \textit{Balloons} and \textit{Office}. Image courtesy of Kede ma.}
\label{fig_shen11}
\end{figure*}
\vspace{-0.2cm}
\section{Experiments and Results}
\label{sec:exp_results}
\vspace{-0.2cm}
We have conducted extensive evaluation and comparison study against state-of-the-art algorithms for variety of natural images. For evaluation, we have chosen standard image sequences to cover different image characteristics including indoor and outdoor, day and night, natural and artificial lighting, linear and non-linear exposure. The proposed algorithm is compared against seven best performing MEF algorithms, (1) Mertens09 \cite{mertens2007exposure}, (2) Li13 \cite{shutao13} (3) Li12 \cite{shutao12} (4) Ma15 \cite{ma2015multi} (5) Raman11 \cite{Shanmuga2011} (6) Shen11 \cite{shen2011generalized} and (7) Guo17 \cite{zhengguo2017detail}. In order to evaluate the performance of algorithms objectively, we adopt MEF SSIM. Although number of other IQA models for general image fusion have also been reported, none of them makes adequate quality predictions of subjective opinions \cite{ma2015perceptual}. 
\vspace{-0.2cm}
\subsection{DeepFuse - Baseline} 
\vspace{-0.2cm}
So far, we have discussed on training CNN model in unsupervised manner. One interesting variant of that would be to train the CNN model with results of other state-of-art methods as ground truth. This experiment can test the capability of CNN to learn complex fusion rules from data itself without the help of MEF SSIM loss function. The ground truth is selected as best of Mertens \cite{mertens2007exposure} and GFF \cite{shutao13} methods based on MEF SSIM score\footnote{In a user survey conducted by Ma \textit{et al.} \cite{ma2015perceptual}, Mertens and GFF results are ranked better than other MEF algorithms}. The choice of loss function to calculate error between ground truth and estimated output is very crucial for training a CNN in supervised fashion. The Mean Square Error or $\ell_2$ loss function is generally chosen as default cost function for training CNN. The $\ell_2$ cost function is desired for its smooth optimization properties. While $\ell_2$ loss function is better suited for classification tasks, they may not be a correct choice for image processing tasks \cite{zhao2015loss}. It is also a well known phenomena that MSE does not correlate well with human perception of image quality \cite{wang2004image}. In order to obtain visually pleasing result, the loss function should be well correlated with HVS, like Structural Similarity Index (SSIM) \cite{wang2004image}.  We have experimented with different loss functions such as $\ell_1$, $\ell_2$ and SSIM.

\begin{figure}[th]
\centering
\subfloat[Underexposed image]{\includegraphics[width=1in]{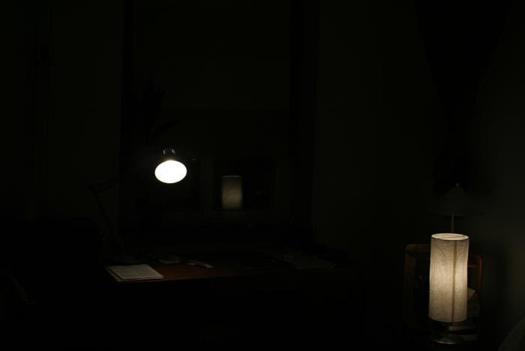}%
\label{fig1_ue_case}}
\hspace{0.05cm}
\subfloat[Ma \emph{et al.} \cite{ma2015multi}]{\includegraphics[width=1in]{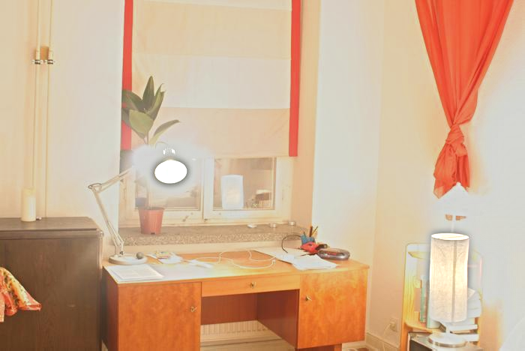}%
\label{fig1_ue_case}}
\hspace{0.05cm}
\subfloat[Zoomed result of (b)]{\includegraphics[width=0.72in]{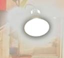}%
\label{fig1_s27_ma}}
\\	
\subfloat[Overexposed image]{\includegraphics[width=1in]{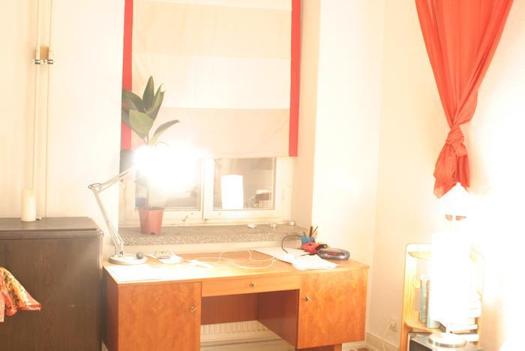}%
\label{fig1_oe_case}}
\hspace{0.05cm}
\subfloat[DF - Unsupervised]{\includegraphics[width=1in]{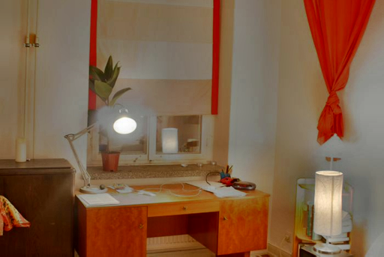}%
\label{fig1_oe_case}}
\hspace{0.05cm}
\subfloat[Zoomed result of (e)]{\includegraphics[width=0.7in]{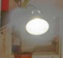}%
\label{fig1_s27_df}}
\caption{Comparison of the proposed method with Ma \emph{et al.} \cite{ma2015multi} for \textit{Table} sequence. The zoomed region of result by Ma \emph{et al.} \cite{ma2015multi} shows the artificial halo artifact effect around edges of lamp. Image courtesy of Kede ma.}
\label{fig_seq27}
\end{figure}
The fused image appear blurred when the CNN was trained with $\ell_2$ loss function. This effect termed as \textit{regression to mean}, is due to the fact that $\ell_2$ loss function compares the result and ground truth in a pixel by pixel manner. The result by $\ell_1$ loss gives sharper result than $\ell_2$ loss but it has halo effect along the edges. Unlike $\ell_1$ and $\ell_2$, results by CNN trained with SSIM loss function are both sharp and artifact-free. Therefore, SSIM is used as loss function to calculate error between generated output and ground truth in this experiment. 

The quantitative comparison between DeepFuse baseline and unsupervised method is shown in Table \ref{tab:mef_ssim}. The MEF SSIM scores in Table \ref{tab:mef_ssim} shows the superior performance of DeepFuse unsupervised over baseline method in almost all test sequences. The reason is due to the fact that for baseline method, the amount of learning is upper bound by the other algorithms, as the ground truth for baseline method is from Merterns \emph{et al.} \cite{mertens2007exposure} or Li \emph{et al.} \cite{shutao13}. We see from Table \ref{tab:mef_ssim} that the baseline method does not exceed both of them.

The idea behind this experiment is to combine advantages of all previous methods, at the same time avoid shortcomings of each. From Fig. \ref{fig:lighthouse}, we can observe that though DF-baseline is trained with results of other methods, it can produce results that do not have any artifacts observed in other results.
\begin{figure}[h]
\centering
\subfloat[Ma \emph{et al.} \cite{ma2015multi}]{\includegraphics[width=1in]{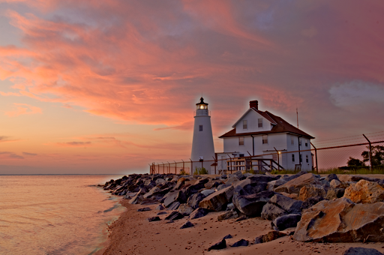}%
\label{fig1_ue_case}}
\hspace{0.005cm}
\subfloat[Zoomed result of (a)]{\includegraphics[width=1.45in]{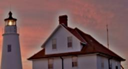}%
\label{fig1_lgt_ma}}
\\	
\subfloat[DF - Unsupervised]{\includegraphics[width=1in]{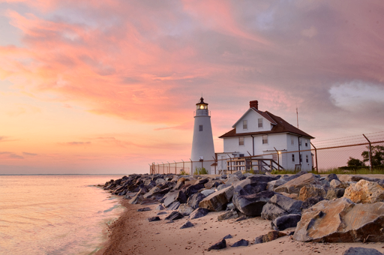}%
\label{fig1_oe_case}}
\hspace{0.005cm}
\subfloat[Zoomed result of (c)]{\includegraphics[width=1.43in]{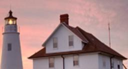}%
\label{fig1_lgt_df}}
\caption{Comparison of the proposed method with Ma \emph{et al.} \cite{ma2015multi}. A close-up look on the results for \textit{Lighthouse} sequence. The results by Ma \emph{et al.} \cite{ma2015multi} show a halo effect along the roof and lighthouse. Image courtesy of Kede Ma.}
\label{fig_lighthouse}
\end{figure}
\vspace{-0.2cm}
\subsection{Comparison with State-of-the-art}
\vspace{-0.2cm}
\textit{Comparison with Mertens} \emph{et al.}: Mertens \emph{et al.} \cite{mertens2007exposure} is a simple and effective weighting based image fusion technique with multi resolution blending to produce smooth results. However, it suffers from following shortcomings: (a) it picks ``best" parts of each image for fusion using hand crafted features like saturation and well-exposedness. This approach would work better for image stacks with many exposure images. But for exposure image pairs, it fails to maintain uniform brightness across whole image. Compared to Mertens \emph{et al.}, DeepFuse produces images with consistent and uniform brightness across whole image. (b) Mertens \emph{et al.} does not preserve complete image details from under exposed image. In Fig. \ref{fig_hostel1}(d), the details of the tile area is missing in Mertens \emph{et al.}'s result. The same is the case in Fig. \ref{fig_hostel1}(j), the fine details of the lamp are not present in the Mertens \emph{et al.} result. Whereas, DeepFuse has learned filters that extract features like edges and textures in C1 and C2, and preserves finer structural details of the scene. 

\textit{Comparison with Li} \emph{et al.} \cite{shutao12} \cite{shutao13}: It can be noted that, similar to Mertens \emph{et al.} \cite{mertens2007exposure}, Li \emph{et al.} \cite{shutao12} \cite{shutao13} also suffers from non-uniform brightness artifact (Fig. \ref{fig_shen11}). In contrast, our algorithm provides a more pleasing image with clear texture details. 

\textit{Comparison with Shen} \emph{et al.} \cite{shen2011generalized}: The results generated by Shen \emph{et al.} show contrast loss and non-uniform brightness distortions (Fig. \ref{fig_shen11}). In Fig. \ref{fig_shen11}(e1), the brightness distortion is present in the cloud region. The cloud regions in between balloons appear darker compared to other regions. This distortion can be observed in other test images as well in Fig. \ref{fig_shen11}(e2). However, the DeepFuse (Fig. \ref{fig_shen11}(f1) and (f2) ) have learnt to produce results without any of these artifacts.

\textit{Comparison with Ma} \emph{et al.} \cite{ma2015multi}: Fig. \ref{fig_seq27} and \ref{fig_lighthouse} shows comparison between results of Ma \emph{et al.} and DeepFuse for Lighthouse and Table sequences. Ma \emph{et al.} proposed a patch based fusion algorithm that fuses patches from input images based on their patch strength. The patch strength is calculated using a power weighting function on each patch. This method of weighting would introduce unpleasant halo effect along edges (see Fig. \ref{fig_seq27} and \ref{fig_lighthouse}). 

\textit{Comparison with Raman} \emph{et al.} \cite{Shanmuga2011}: Fig. \ref{fig:lighthouse}(f) shows the fused result by Raman \emph{et al.} for House sequence. The result exhibit color distortion and contrast loss. In contrast, proposed method produces result with vivid color quality and better contrast.

After examining the results by both subjective and objective evaluations, we observed that our method is able to faithfully reproduce all the features in the input pair. We also notice that the results obtained by DeepFuse are free of artifacts such as darker regions and mismatched colors. Our approach preserves the finer image details along with higher contrast and vivid colors. The quantitative comparison between proposed method and existing approaches in Table \ref{tab:mef_ssim} also shows that proposed method outperforms others in most of the test sequences. From the execution times shown in Table \ref{tab:comp_time} we can observe that our method is roughly 3-4$\times$ faster than Mertens \emph{et al}. DeepFuse can be easily extended to more input images by adding additional streams before merge layer. We have trained DeepFuse for sequences with 3 and 4 images. For sequences with 3 images, average MEF SSIM score for DF is 0.987 and 0.979 for Mertens \textit{et al}. For sequences with 4 images, average MEF SSIM score for DF is 0.972 and 0.978 for Mertens \textit{et al.} For sequences with 4 images, we attribute dip in performance to insufficient training data. With more training data, DF can be trained to perform better in such cases as well. 
\begin{figure}[t]
\centering
\subfloat{\includegraphics[width=1.05in]{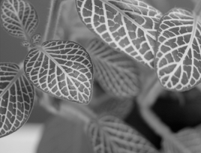}%
\label{fig1_ue_case}}
\hspace{0.02cm}
\subfloat{\includegraphics[width=1.05in]{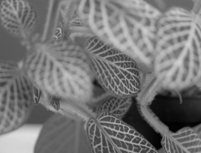}%
\label{fig1_ue_case}}
\hspace{0.02cm}
\subfloat{\includegraphics[width=1.05in]{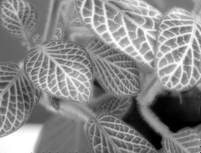}%
\label{fig1_ue_case}}
\hspace{0.02cm}
\clearsubcaptcounter
\\\vspace{-1.5mm}
\subfloat[Near focused image]{\includegraphics[width=1.05in]{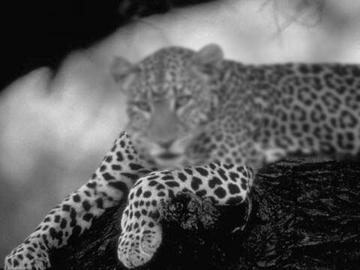}%
\label{fig1_s24_mer}}
\hspace{0.02cm}
\subfloat[Far focused image]{\includegraphics[width=1.05in]{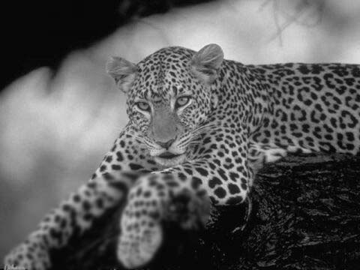}%
\label{fig1_s24_mer}}
\hspace{0.02cm}
\subfloat[DF result]{\includegraphics[width=1.05in]{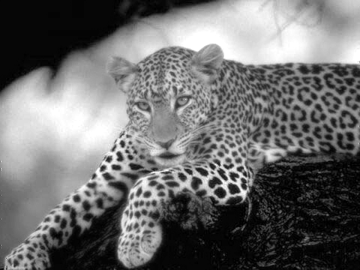}%
\label{fig1_fmmr_case}}
\hspace{0.02cm}
\caption{Application of DeepFuse CNN to multi-focus fusion. The first two column images are input varying focus images. The All-in-focus result by DeepFuse is shown in third column. Images courtesy of Liu \emph{et al.} \cite{liu2015multi}. Image courtesy of Slavica savic.}
\label{fig_focus}
\end{figure}
\subsection{Application to Multi-Focus Fusion}
In this section, we discuss the possibility of applying our DeepFuse model for solving other image fusion problems. Due to the limited depth-of-field in the present day cameras, only object in limited range of depth are focused and the remaining regions appear blurry. In such scenario, Multi-Focus Fusion (MFF) techniques are used to fuse images taken with varying focus to generate a single all-in-focus image. MFF problem is very similar to MEF, except that the input images have varying focus than varying exposure for MEF. To test the generalizability of CNN, we have used the already trained DeepFuse CNN to fuse multi-focus images without any fine-tuning for MFF problem. Fig. \ref{fig_focus} shows that the DeepFuse results on publicly available multi-focus dataset show that the filters of CNN have learnt to identify proper regions in each input image and successfully fuse them together. It can also be seen that the learnt CNN filters are generic and could be applied for general image fusion.

\begin{table}[t]
\centering
\caption{\textbf{Computation time}: Running time in seconds of different algorithms on a pair of images. The numbers in bold denote the least amount of time taken to fuse. $\ddagger$: tested with NVIDIA Tesla K20c GPU, $\dagger$: tested with Intel\rr Xeon @ 3.50 GHz CPU}
\label{tab:comp_time}
\begin{tabular}{@{}rcccc@{}}
\toprule
\multicolumn{1}{c}{Image size} & Ma$15^\dagger$ & Li$13^\dagger$ & Mertens$07^\dagger$ & $DF^\ddagger$ \\ \midrule
                 512*384       & 2.62   & 0.58   & 0.28        & \textbf{0.07}     \\
                 1024*768      & 9.57   & 2.30   & 0.96        & \textbf{0.28}      \\
                 1280*1024     & 14.72  & 3.67   & 1.60        & \textbf{0.46}      \\
                 1920*1200     & 27.32  & 6.60   & 2.76        & \textbf{0.82}      \\ \bottomrule
\end{tabular}

\end{table}
\section{Conclusion and Future work}
\label{sec:typestyle}
In this paper, we have proposed a method to efficiently fuse a pair of images with varied exposure levels to produce an output which is artifact-free and perceptually pleasing. DeepFuse is the first ever unsupervised deep learning method to perform static MEF. The proposed model extracts set of common low-level features from each input images. Feature pairs of all input images are fused into a single feature by merge layer. Finally, the fused features are input to reconstruction layers to get the final fused image. We train and test our model with a huge set of exposure stacks captured with diverse settings. Furthermore, our model is free of parameter fine-tuning for varying input conditions. Finally, from extensive quantitative and qualitative evaluation, we demonstrate that the proposed architecture performs better than state-of-the-art approaches for a wide range of input scenarios. 
\par In summary, the advantages offered by DF are as follows: 1) Better fusion quality: produces better fusion result even for extreme exposure image pairs, 2) SSIM over $\ell_1$ : In \cite{zhao2015loss}, the authors report that $\ell_1$ loss outperforms SSIM loss function. In their work, the authors have implemented approximate version of SSIM and found it to perform sub-par compared to $\ell_1$. We have implemented the exact SSIM formulation and observed that SSIM loss function perform much better than MSE and $\ell_1$. Further, we have shown that a complex perceptual loss such as MEF SSIM can be successfully incorporated with CNNs in absense of ground truth data. The results encourage the research community to examine other perceptual quality metrics and use them as loss functions to train a neural net. 3) Generalizability to other fusion tasks: The proposed fusion is generic in nature and could be easily adapted to other fusion problems as well. In our current work, DF is trained to fuse static images. For future research, we aim to generalize DeepFuse to fuse images with object motion as well. 
{\small
\bibliographystyle{ieee}
\bibliography{egbib}
}

\end{document}